\documentclass{ISMA_USD}

\usepackage{subcaption}

\usepackage{tikz}
\usetikzlibrary{shapes, arrows, positioning, fit, backgrounds}
\usetikzlibrary{intersections}
\usetikzlibrary{decorations, decorations.markings}
\usetikzlibrary{patterns,snakes, calc}

\tikzstyle{ground}=[preaction={fill,top color=black!10,bottom color=black!5,shading angle=25},
                    fill,pattern=north east lines,draw=none,minimum width=0.3,minimum height=0.6]

\tikzstyle{mass}=[line width=0.6,black!30!black,fill=red!40!black!10,rounded corners=1,
                  top color=green!40!black!20,bottom color=green!40!black!10,shading angle=20]

\tikzstyle{spring}=[line width=0.8,blue!7!black!80,snake=coil,segment amplitude=5,segment length=5,line cap=round]

\tikzstyle{damper}=[thick,decoration={markings,  
   mark connection node=dmp,
   mark=at position 0.5 with 
   {
     \node (dmp) [thick,inner sep=0pt,transform shape,rotate=-90,minimum
 width=15pt,minimum height=3pt,draw=none] {};
     \draw [thick] ($(dmp.north east)+(2pt,0)$) -- (dmp.south east) -- (dmp.south
 west) -- ($(dmp.north west)+(2pt,0)$);
     \draw [thick] ($(dmp.north)+(0,-5pt)$) -- ($(dmp.north)+(0,5pt)$);
   }
 }, decorate]

\usepackage{xcolor, soul}
\sethlcolor{yellow}
\usepackage{comment}

\hypersetup{pdftitle  = {Regularising NARX models by MTL},
	pdfauthor = {S.C. Bee, L.A. Bull, N. Dervilis, K. Worden},
	pdfkeywords = {NARX, Multi-task Learning}}

\title{Regularising NARX models with multi-task learning}

\author[1]{S.C.\ Bee}
\author[2]{L.A.\ Bull}
\author[1]{N.\ Dervilis}
\author[1]{K.\ Worden}
	
\affil[1]{Dynamics Research Group, Department of Mechanical Engineering, University of Sheffield, \NewLineAffil Mappin Street, Sheffield S1 3JD, UK}
	\affil[2]{Department of Engineering, University of Cambridge, \NewLineAffil Cambridge, United Kingdom, UK, CB3 0FA}

\date{}

\begin{document}
%\bstctlcite{IEEEtran:BSTadapt}	

%%%%%%%%%%%%
% Abstract %
%%%%%%%%%%%%

% Enter your abstract in the command below.

\abstract{
A Nonlinear Auto-Regressive with eXogenous inputs (NARX) model can be used to describe time-varying processes; where the output depends on both previous outputs and current/previous external input variables. One limitation of NARX models is their propensity to overfit and result in poor generalisation for future predictions. The proposed method to help to overcome the issue of overfitting is a NARX model which predicts outputs at both the current time and several lead times into the future. This is a form of multi-task learner (MTL); whereby the lead time outputs will regularise the current time output. This work shows that for high noise level, MTL can be used to regularise NARX with a lower Normalised Mean Square Error (NMSE) compared to the NMSE of the independent learner counterpart.
}

\maketitle

%%%%%%%%%%%%%%%%%%%%%%%%%%
% Beginning of the paper %
%%%%%%%%%%%%%%%%%%%%%%%%%%

\section{Introduction}

Auto-regressive neural networks utilise previous outputs to predict a current output. Nonlinear Auto-Regressive with eXogenous input (NARX) neural networks (NN) are a type of auto-regressive model which include both  previous outputs to predict a current output, as well as external input variables \cite{Urbani1994}. NARX models have many applications for SHM systems, for example the prediction of damage-sensitive features in a live system.  

\begin{figure}[h]
  \centering
  \begin{minipage}{.5\textwidth}
      \centering

      \tikzstyle{nodeI} = [thick,draw=blue,fill=blue!20,circle,minimum size=45]
      \tikzstyle{nodeH} = [thick,draw=red,fill=red!20,circle,minimum size=45]
      \tikzstyle{nodeO} = [thick,draw=green,fill=green!20,circle,minimum size=45]
      \tikzstyle{line} = [draw, -latex']
      % First TikZ figure
      \begin{tikzpicture}[thick,scale=0.6, every node/.style={scale=0.6}]
          \node [nodeI] (input1) at (1,7) {$y_{t-1}$};
          \node [nodeI] (input2) at (1, 4.5) {$y_{t-m-1}$};
          \node [nodeI] (input3) at (1, 2.5) {$x_{t}$};
          \node [nodeI] (input4) at (1, 0) {$x_{t-m}$};
          
          \node [nodeH] (hidden1) at (4,5) {};
          \node [nodeH] (hidden2) at (4,2) {};
      
          \node [nodeO] (output1) at (7,3.5) {$y_{t}$};
      
          % Draw edges
          \path [line] (input1) -> (hidden1);
          \path [line] (input2) -> (hidden1);
          \path [line] (input3) -> (hidden1);
          \path [line] (input4) -> (hidden1);
      
          \path [line] (input1) -> (hidden2);
          \path [line] (input2) -> (hidden2);
          \path [line] (input3) -> (hidden2);
          \path [line] (input4) -> (hidden2);
      
          \path [line] (hidden1) -> (output1);
          \path [line] (hidden2) -> (output1);
      
          % Draw ellipsis
          \path (input1) -- (input2) node [black, font=\large, midway, sloped] {$\dots$};
          
          \path (input3) -- (input4) node [black, font=\large, midway, sloped] {$\dots$};

            \path (hidden1) -- (hidden2) node [black, font=\large, midway, sloped] {$\dots$};
    
      \end{tikzpicture}

  \end{minipage}%
  \begin{minipage}{.5\textwidth}
      \centering

          \tikzstyle{nodeI} = [thick,draw=blue,fill=blue!20,circle,minimum size=45]
\tikzstyle{nodeH} = [thick,draw=red,fill=red!20,circle,minimum size=45]
\tikzstyle{nodeO} = [thick,draw=green,fill=green!20,circle,minimum size=45]
\tikzstyle{line} = [draw, -latex']
      % Second TikZ figure
      \begin{tikzpicture}[thick,scale=0.6, every node/.style={scale=0.6}]
          \node [nodeI] (input1) at (1,7) {$y_{t-1}$};
          \node [nodeI] (input2) at (1, 4.5) {$y_{t-m-1}$};
          \node [nodeI] (input3) at (1, 2.5) {$x_{t}$};
          \node [nodeI] (input4) at (1, 0) {$x_{t-m}$};
          
          \node [nodeH] (hidden1) at (4,5) {};
          \node [nodeH] (hidden2) at (4,2) {};
      
          \node [nodeO] (output1) at (7,5) {$y_{t}$};
          \node [nodeO] (output2) at (7,2) {$y_{t+m}$};
      
          % Draw edges
          \path [line] (input1) -> (hidden1);
          \path [line] (input2) -> (hidden1);
          \path [line] (input3) -> (hidden1);
          \path [line] (input4) -> (hidden1);
      
          \path [line] (input1) -> (hidden2);
          \path [line] (input2) -> (hidden2);
          \path [line] (input3) -> (hidden2);
          \path [line] (input4) -> (hidden2);
      
          \path [line] (hidden1) -> (output1);
          \path [line] (hidden2) -> (output1);
          \path [line] (hidden1) -> (output2);
          \path [line] (hidden2) -> (output2);
      
          % Draw ellipsis
          \path (input1) -- (input2) node [black, font=\large, midway, sloped] {$\dots$};
          
          \path (input3) -- (input4) node [black, font=\large, midway, sloped] {$\dots$};

          \path (hidden1) -- (hidden2) node [black, font=\large, midway, sloped] {$\dots$};

          \path (output1) -- (output2) node [black, font=\large, midway, sloped] {$\dots$};

      \end{tikzpicture}

  \end{minipage}

  \caption{Example of a standard NARX neural network, \emph{left}, and example of a MT-NARX neural network, \emph{right}.}
  \label{fig:NN}
\end{figure}
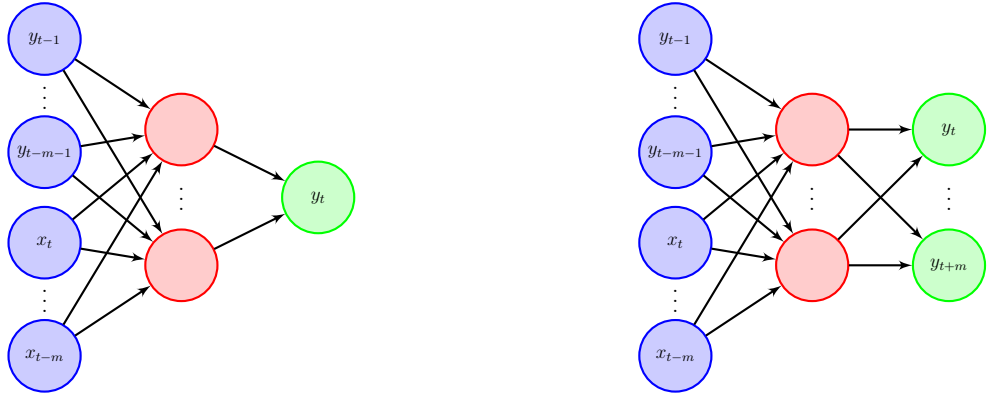

Figure \ref{fig:NN} \emph{left} depicts an example of a standard (ST) NARX NN (ST-NARX). To determine the response of the system ($y_t$) previous responses (\emph{lags}) are utilised. The number of lags ($m$) is a hyper-parameter of the model. In addition to the response, the input forcing ($x$) for the current timestamp ($t$) and $m-1$ lags are inputs to the model.

Each consecutive output prediction from the model will become an input into the next iteration to the model. One of the pitfalls of the NARX NN is that, when predicting multiple steps ahead, errors from the model can propagate through to future predictions. However, if the model generalises well, one can consider such that the physics of the system are sufficiently approximated \cite{Worden2007}, and the model should be able to make good predictions.One limitation of NARX models is their propensity to overfit and result in poor generalisation for future predictions. This work suggests that poor generalisation of NARX NN \cite{Tu2023} could be overcome by utilising \emph{multi-task learning}.

Multi-task learning (MTL) is a suite of algorithms which learn tasks simultaneously, as opposed to independently from one another. In recent years there has been an increase in the research into MTL for structural datasets \cite{Hernández-González2024, Shu2023, Wan2022, Xue2023}. The closest case of utilising NARX technology for MTL, although not relating to structural datasets, is Nikentari \textit{et. al.} \cite{Nikentari2022}, who successfully combined a Nonlinear AutoRegressive Moving Average with eXogenous input (NARMAX) model and Long-Short Term Memory (LSTM) Neural Networks to form a MT-NARMAX-LSTM which utilised related tasks to reduce over-fitting.

It is important to note that there are several types of MTL which may be applicable to structural datasets \cite{Bee2024}. The recent research into MTL cited all fit into the category of the \emph{natural occurrence of multiple tasks} \cite{Bee2024}, \textit{i.e.} the rationale for research is that the performance of the algorithms improves with the addition of related tasks. The current research suggests that performance could be improved with an MT-NARX which utilises \emph{non-operational features} to improve performance \textit{i.e.} features that are available in training and not testing.

Caruana \cite{caruana1997}, hypothesised that non-operational features could be additional tasks that improve the performance of the main task. The additional outputs influence the loss function and hence the weights during training of the NN, and so the output will have an impact on the model, hence, it can be re-interpreted as an input to the model. 

For time-series prediction, future responses (leads), $i.e.$ $y_{t+1}, y_{t+2}, ..., y_{t+i}$ where $i\in\mathbb{N}$, can be utilised as non-operational features. Multiple non-operational features increase the model's complexity, hence the number of computations required. By creating a symmetry of lags and leads, the predicted value sits in the middle of the outputs so the MTL approach can be interpreted as a smoothing methodology, akin to a simple moving average. Figure \ref{fig:NN} \emph{right}, shows a NARX network with $m$ additional non-operational features, this is an example of a multi-task NARX (MT-NARX).

%%The closest case of utilising NARX technology for MTL is Nikentari \textit{et. al.} \cite{Nikentari2022}, who successfully combined Nonlinear AutoRegressive Moving Average with eXogenous input (NARMAX) model and Long-Short Term Memory (LSTM) Neural Networks to form a MT-NARMAX-LSTM which utilised related tasks to reduce over-fitting. Although this is using similar technology, this is an example of the natural occurrence of multiple tasks \cite{Bee2024}, rather than the use of non-operational features. 

\section{Development of case study}

\subsection{Base dataset}
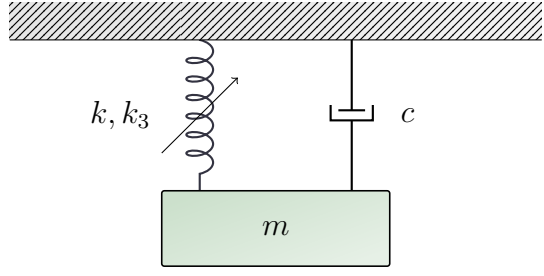
\begin{figure}[ht]
  \centering
  
  % VERTICAL spring - extension at rest
  \begin{tikzpicture}
    \def\H{0.5} % ceiling height
    \def\W{7.0}  % ceiling width
    \def\h{1}  % mass height
    \def\w{3}  % mass width
    \def\y{2}  % mass height from wall
  
    \draw[spring,segment length=7] (-1,0) -- (-1,-\y);
    \draw[ground] (-\W/2,0) rectangle++ (\W,\H);
    \draw (-\W/2,0) --++ (\W,0);
    \draw[mass] (-\w/2,-\y) rectangle++ (\w,-\h) node[midway] {$m$};
  
    \draw[damper] (1,-\y) -- (1,0);
  
    \node[anchor=west] at (1.5, -\y/2) {$c$};
    
    \draw[->] (-1.5,-1.5) -- (-0.5,-0.5);
    \node[anchor=east] at (-1.5, -\y/2) {$k, k_3$};
  
  \end{tikzpicture}
  
  \caption{SDOF Duffing oscillator with linear and cubic stiffness $k$ and $k_3$, respectively, mass $m$ and damping coefficient $c$.}
  \label{fig:Duffing}
  \end{figure}

To assess the benefits of MT-NARX, a single-degree-of-freedom (SDOF) Duffing oscillator is presented as a nonlinear example. Figure \ref{fig:Duffing} shows a SDOF Duffing oscillator, the system can be can be represented by the following equation of motion:  
  \begin{equation}
    \label{eq:motion}
    m\ddot{y} + c \dot{y} + ky + k_3y^{3} = x(t)
  \end{equation}
  
where $\quad\dot{}\quad$ and $\quad\ddot{}\quad$ denote the first and second order differentiation with respect to time, respectively.\\

The parameters for the oscillator were selected to align with previous works \cite{Champneys2022}; $m=1$ kg, $c=20$ Ns/m, $k = 10^4$ N/m and $k_3 = 5 \times 10^9$ N/m\textsuperscript{3}. The cubic stiffness, $k_3$, represents the nonlinearity of the system and, akin to the other parameters, is fixed. As $k_3>0$, the system has a hardening characteristic \cite{Worden2019}. The larger the excitation, the greater the extent of nonlinearities that the system will exhibit, as such, the system was excited with Gaussian white noise, $x(t) \sim N(0, 10)$ so that the nonlinear behaviour would be present. 

The simulated response of the system here was stepped through time using the fourth-order Runge-Kutta. A Butterworth filter \cite{Butterworth1930} was applied to limit the excitation frequencies to 50Hz, a sampling frequency of 500Hz is applied. 

Transient behaviour, which may be present from the initial conditions, was removed by discarding the first $10^4$ samples from the dataset and keeping only the final 4000 data points for analysis.

\subsection{Representation of noise}
The level of noise present in a dataset will impact the results of a predictive model. For a NARX-NN, there are two sensor readings, the input force, $x(t)$ and the displacement, $y$ and therefore, two places where noise can be introduced. There are three noise-cases which can be introduced to equation (\ref{eq:motion}): 
\begin{itemize}
  \item Trial 1: On the output only: $m\ddot{(y + \epsilon_y)} + c \dot{(y + \epsilon_y)} + k(y+ \epsilon_y) + k_3(y + \epsilon_y)^{3} = x(t)$. 
  \item Trial 2: On the input only: $m\ddot{y} + c \dot{y} + ky + k_3y^{3} = x(t) + \epsilon_x$.
  \item Trial 3: On both the input and the output: $m\ddot{(y + \epsilon_y)} + c \dot{(y + \epsilon_y)} + k(y+ \epsilon_y) + k_3(y + \epsilon_y)^{3} = x(t) + \epsilon_x$.
\end{itemize}

\begin{figure}[ht]
  \centering
  \includegraphics[scale = 0.7]{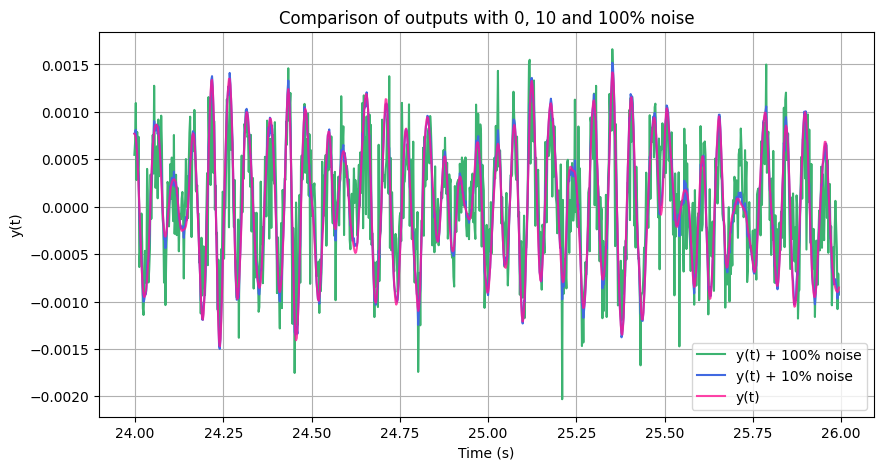}
  \caption{The system response over time with no noise (\emph{pink}), 10\% noise (minor noise, \emph{blue}), and high noise (\emph{green}).}
  \label{fig:noise}
\end{figure}

For each of the three noise-cases, noise, $a$, is introduced. The noise is applied as a Gaussian distribution and added to the simulated time-series. For noise applied to the force, $\epsilon_x \sim N(0, a RMS)$, and, for noise applied to the displacement, $\epsilon_y \sim N(0, a \sigma _y)$, where $a$ refers to the fraction of noise, $\{0, 0.1, 0.3, 0.5, 1\}$. For the third noise-case, $a$ is equal for both $RMS$ and $\sigma _y^2$. Noise was trialled at the following levels: 1\%, 10\%, 30\%, 100\% and 150\%. Figure \ref{fig:noise} shows the difference between no, low and high noise levels on the output parameter. As the noise is a Gaussian distribution, even in the high noise cases, the underlying trend of the data is somewhat visible.

\subsection{NN development}

The generated dataset consisted of 4000 samples of input and displacement data. A data split of $2:1:1$ was used for training, validation and testing, respectively. The training dataset represents half of the dataset. 

The network was designed such that there was one hidden layer using the hyperbolic tangent (tanh) function for activation. Preliminary investigations reviewed the use of hyperbolic tangent, Sigmoid and Huber. Similar good performances were seen from both tanh and Sigmoid, however, the tanh was selected to align with previous research \cite{Chance1998}. A bias unit was added to both the hidden layer and the output layer. 

The Adam optimisation algorithm was used to optimise the learning rate throughout the network training. The loss function used was mean-squared error, 
\begin{equation}
    MSE(\hat{y}) = \frac{1}{N}\sum^N_{i=1} (\hat{y}_i - y_i)^2
    \label{eq:MSE}
\end{equation}

Early stopping was utilised with a high patience of 100 epochs. 

\subsection{Performance metrics}
There are two outputs from the model which can be assessed. The one-step ahead prediction (OSA) error from the model is synonymous with the loss which will be analysed during the model training. OSA error is used to rank the networks during training. However, a strong OSA does not necessarily correspond to a strong model-predicted output (MPO). MPO feeds the model predictions back as inputs into the model for the next prediction; hence, the MPO is more likely to demonstrate the dynamics which have been learnt by the model.

The fit of the trained network (OSA and MPO) is assessed using the normalised mean square error: 
\begin{equation}
  NMSE(\hat{y}) = \frac{100}{N\sigma^2}\sum_{i=1}^{N}(y_i - \hat{y}_i)^2
\end{equation}

\noindent The performance of models is categorised such that less than 5\% is a good fit and less than 1\% is an excellent fit. 

\subsection{Hyper-parameter training}
For each of the three trials the training began with the ST-NARX. A NN for each combination of the number of lags and the number of nodes was trained for 10 random initialisation states (shown in Table (\ref{table:ST-training})). This provided a gauge as to the range of lags and nodes which should be ran in the final hyperparameter training. 

\begin{table}[ht]
\centering
\begin{tabular}{c|cc}
$ Trial* $   & $N_{lags} $ & $N_{nodes}$ \\
\hline
$1$  &    $\{3, 4, ..., 15\}$        &    $\{2, 3, ..., 30\}$       \\
$2$  &  $\{3, 4, ..., 10\}$          &    $\{2, 3, ..., 20\}$       \\
$3$  &    $\{3, 4, ..., 20\}$        &    $\{2, 3, ..., 30\}$       \\
\end{tabular}
\caption{Hyperparameters for ST-NARX training - * 150\% noise case had up to 50 nodes to account for the additional requirements of the model}
\label{table:ST-training}
\end{table}

The networks were then trained with a narrower range of hyperparameters for 10 iterations, this time the optimal network was selected as the network with the lowest NMSE MPO on the validation dataset. 

For comparison, the number of lags for the optimal ST-NARX was then set as the number of lags for the MT-NARX. The number of leads that the MT-NARX was trained on was $N_{leads} \in \{N_{lags}, N_{lags + 1}, ..., N_{lags + 3} \}$. The number of units was also optimised in a similar vein to the ST-NARX. The number of nodes was increased for increasing level of noise, Table (\ref{table:MT-training}) shows the parameters used for the initial training. 

\begin{table}[ht]
\centering
\begin{tabular}{c|cc}
Noise level   & Trial 1 / 3 & Trial 2 \\
\hline
$1\%$  &    $\{10, 11, ..., 50\}$        &    $\{2, 3, ..., 15\}$       \\
$10\%$  &  $\{25, 26, ..., 50\}$          &    $\{2, 3, ..., 15\}$       \\
$30\% $  &    $\{25, 26, ..., 60\}$        &    $\{2, 3, ..., 20\}$       \\
$50\% $  &    $\{25, 26, ..., 60\}$        &    $\{2, 3 ..., 20\}$       \\
$100\% $  &    $\{25, 26, ..., 70\}$        &    $\{10, ..., 30\}$       \\
$150\% $  &    $\{40, 41, ..., 80\}$*        &    Not assessed       \\
\end{tabular}
\caption{Number of nodes for MT-NARX training \\
 \emph{\it{* 150\% noise for Trial 1 only}}}
\label{table:MT-training}
\end{table}

When training the NN the algorithm seeks to minimise the MSE (equation (\ref{eq:MSE})). For an MT-NARX, the principle is the same where the total loss across all of the outputs is minimised. However, the optimal solution desired is one which minimises the first output only, therefore, without altering the loss function, the solutions obtained may have a low loss function overall but not necessarily the lowest in the desired parameter. To allow the MT-NARX to be comparable to the ST-NARX, the number of iterations that each architecture of the NN was trained was increased from 10, in the case of the ST-NARX to 100, 200, and 10 iterations for Trials 1, 2 and 3, respectively. The increased number of iterations were performed on a narrower range of hyper parameters, as per ST-NARX methodology. Again, following training, the optimal network was selected as the network with the lowest NMSE MPO error on the validation dataset.

\section{Results}

\begin{figure}[h!]
    \centering
    \begin{subfigure}{0.5\textwidth}
        \centering
        \includegraphics[width=\textwidth]{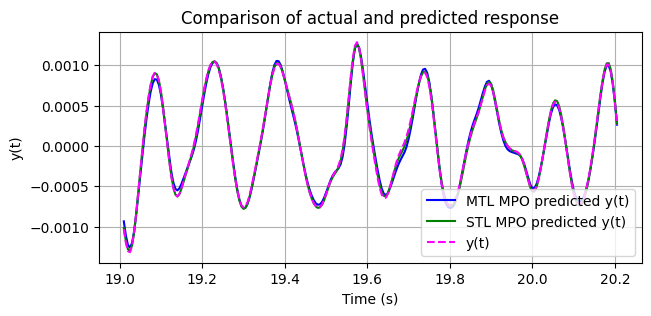}
    \caption{1\% Noise}
    \end{subfigure}%
    \hfill
    \begin{subfigure}{0.5\textwidth}
        \centering
        \includegraphics[width=\textwidth]{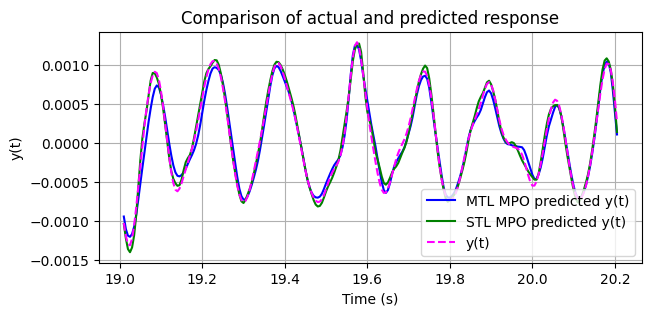}
    \caption{50\% Noise}
    \end{subfigure}%
    \\[0.5cm]
    \begin{subfigure}{0.5\textwidth}
        \centering
        \includegraphics[width=\textwidth]{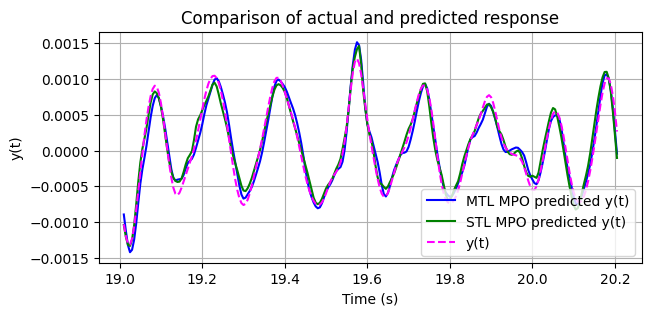}
    \caption{100\% Noise}
    \end{subfigure}%
    \hfill
    \begin{subfigure}{0.5\textwidth}
        \centering
        \includegraphics[width=\textwidth]{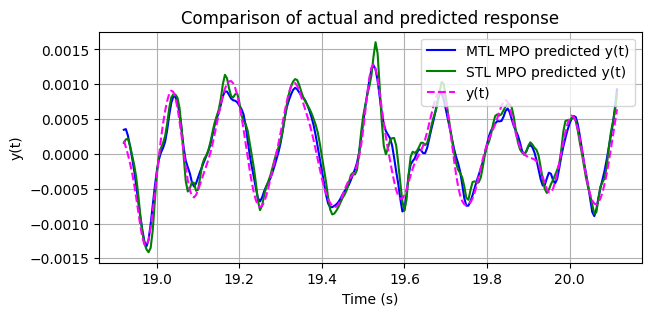}
    \caption{150\% Noise}
    \end{subfigure}%
    \caption{Optimal model for ST-NARX (green) and MT-NARX (blue) for different noise-cases alongside the solution with no noise (pink) for trial 1}
    \label{fig:trial1results}
\end{figure}

Figure (\ref{fig:trial1results}) shows the MPO for the last 250 data points of the test dataset for Trial 1. The ST-NARX MPO and MT-NARX MPO are plotted alongside the simulated output, $y(t)$, without noise added. The fit at low noise is excellent for both models. As expected, with increasing noise the model aligns less with the noiseless data. For the lower levels of noise, ST-NARX outperforms MT-NARX, it is only 100\% noise and 150\% of which MT-NARX outperforms ST-NARX. At the highest level of noise the NMSE is 7.3\% and 6.0\% for ST-NARX and MT-NARX, respectively. 

\begin{figure}[h!]
    \centering
    \begin{subfigure}{0.5\textwidth}
        \centering
        \includegraphics[width=\textwidth]{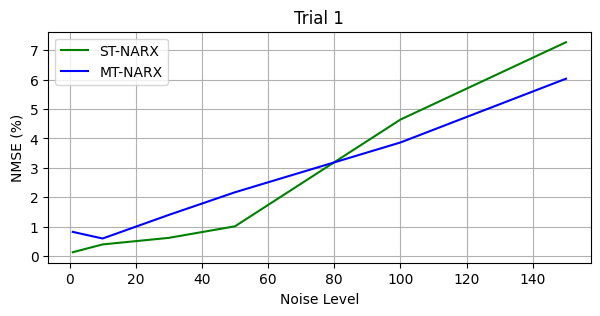}
    \caption{Trial 1 - including 150\% noise level}
    \end{subfigure}%
    \hfill
    \begin{subfigure}{0.5\textwidth}
        \centering
        \includegraphics[width=\textwidth]{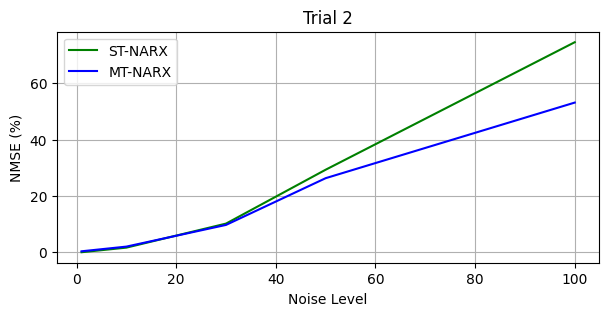}
    \caption{Trial 2}
    \end{subfigure}%
    \\[0.5cm]
    \begin{subfigure}{0.5\textwidth}
        \centering
        \includegraphics[width=\textwidth]{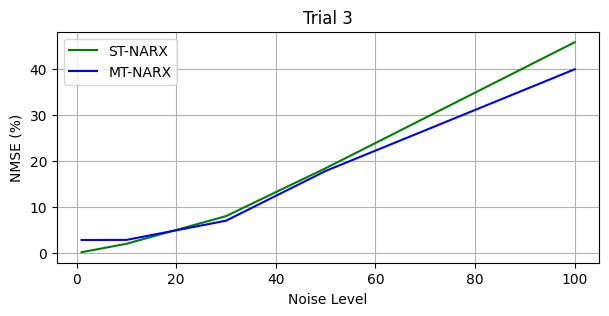}
    \caption{Trial 3}
    \end{subfigure}%
    \hfill
    \caption{NMSE for ST-NARX (green) and MT-NARX (blue) for different noise levels (1\%, 10\%, 30\%, 50\% and 100\%) for the three trials.}
    \label{fig:trialresults}
\end{figure}

A useful relationship is how the NMSE changes with increasing noise for the two models, as shown in Figure (\ref{fig:trialresults}). Although there are only 5 points used in the plots, the ST-NARX has a higher NMSE than the MT-NARX for the higher noise cases for all trials. In Trial 1, the 100\% noise case both of the models are classed as \emph{good} with MT-NARX outperforming ST-NARX. However, when noise is added to the input forcing (Trial 2 and Trial 3), the fit of the model is poor (much greater than 5\%) for noise cases greater than 10\%, though MTL atill offers positive regularisation effects. 

%\pagebreak
\section{Discussion}
\begin{figure}[h!]
    \centering
    \begin{subfigure}{0.5\textwidth}
        \centering
        \centering
        \includegraphics[width=0.9\textwidth]{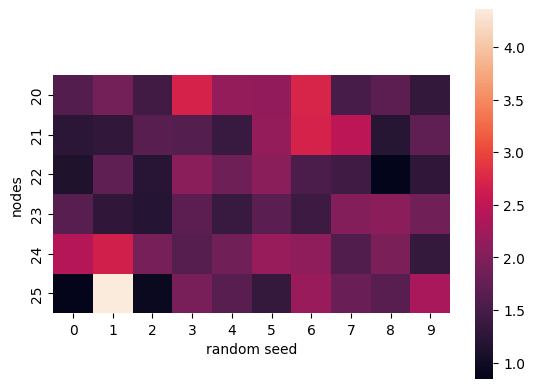}
        \caption{ST-NARX with $15$ lags.}
    \end{subfigure}%
    \hfill
    \begin{subfigure}{0.5\textwidth}
        \centering
        \centering
        \includegraphics[width=0.9\textwidth]{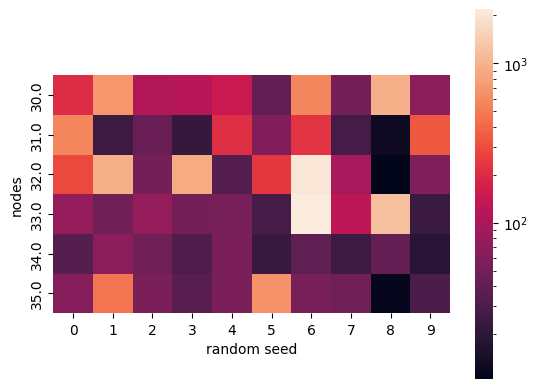}
        \caption{MT-NARX with $15$ lags and $16$ leads with a log scale.}
    \end{subfigure}%
    
    \caption{A Heatmap for the MPO NMSE for different number of nodes and different random initialisation seeds.}
    \label{fig:heatmap}
\end{figure}

Prior to training, the model is initialised with random weights, which are updated by Adam until a minimum is found. However, due to the complexity of the model, it is likely that the optimiser will be trapped in a local minimum. If the architecture had the largest influence on the model then there would be clear trends across different architectures, regardless of the random initialisation; instead, Figure (\ref{fig:heatmap}: $a$) shows that the random seed has as much influence as the number of nodes with one output. Figure (\ref{fig:heatmap}: $b$) demonstrates that the additional outputs in the MT-NARX result in variable MPO NMSE across multiple orders of magnitude. The MPO NMSE is based on the target output only, however, the loss function for the MT-NARX seeks to minimise the loss across \emph{all} of the outputs. A model could reach a local minima with low error for many of the outputs but not necessarily perform well for the intended output, $y_t$, hence, the phenomena observed in the ST-NARX, with regards to local minima, is amplified for the MT-NARX models. 

The complexity of the loss function for the MT-NARX models does not favour optimisation, hence the increase in initialisations. To enable the MT-NARX to be more likely to find a local minima which favours the optimal solution, further work could be done to explore the impact of weightings in the loss function to encourage a solution which favours minimising the loss of the first output. This would require amending equation \ref{eq:MSE} to: 

\begin{equation}
    MSE(\hat{y}) = \frac{1}{N}\sum^N_{i=1} a_i(\hat{y}_i - y_i)^2
    \label{eq:MSE2}
\end{equation}

where $\sum^N_{i=1}a_i = 1$, $a_1 > a_{i+1}$ and for $i>1, a_i = a_{i+1}$.

The higher the weighting that $a_1$ has, the larger the impact the first output (the target output) will have on the loss function. Hence in theory, the more likely the MT-NARX to reach a minima which benefits optimises the target output. 

The results from this piece of work show a small improvement in MPO NMSE with increasing noise. However, in the case of the large noise cases, the model is not a good fit for neither ST-NARX nor MT-NARX. So, although the loss has improved, the model has not improved enough to class it as a good fit. The use of non-operational features in structural problems is still a feasible branch of multi-task learning to explore. However, more motivating results would likely require from further developments to the loss function.

%%%%%%%%%%%%%%%%%%%%
% Acknowledgements %
%%%%%%%%%%%%%%%%%%%%

\section*{Acknowledgements}

The authors acknowledge the support of the UK Engineering and Physical Sciences Research Council (EPSRC) through EP/W005816/1  - Revolutionising Operational Safety and Economy for High-value Infrastructure using Population-based SHM (ROSEHIPS). SCB, ND and KW also acknowledge the support of EPSRC and UK Natural Environment Research Council (NERC) via grant number, EP/S023763/1.

%%%%%%%%%%%%%%
% References %
%%%%%%%%%%%%%%

\bibliography{References}

%%%%%%%%%%%%%%
% Appendices %
%%%%%%%%%%%%%%

% If none, remove the code below.

%\newpage
%\appendix
%\appendixheader
%\section{Checklist}

\end{document}